# $\mathcal{P}$earl: A Probabilistic Chart Parser[*]

**David M. Magerman** and **Mitchell P. Marcus**
CIS Department
University of Pennsylvania
Philadelphia, PA 19104
Internet: magerman@neon.stanford.edu

## Abstract

This paper describes a natural language parsing algorithm for unrestricted text which uses a probability-based scoring function to select the "best" parse of a sentence. The parser, $\mathcal{P}$earl, is a time-asynchronous bottom-up chart parser with Earley-type top-down prediction which pursues the highest-scoring theory in the chart, where the score of a theory represents the extent to which the context of the sentence predicts that interpretation. This parser differs from previous attempts at stochastic parsers in that it uses a richer form of conditional probabilities based on context to predict likelihood. $\mathcal{P}$earl also provides a framework for incorporating the results of previous work in part-of-speech assignment, unknown word models, and other probabilistic models of linguistic features into one parsing tool, interleaving these techniques instead of using the traditional pipeline architecture. In preliminary tests, $\mathcal{P}$earl has been successful at resolving part-of-speech and word (in speech processing) ambiguity, determining categories for unknown words, and selecting correct parses first using a very loosely fitting covering grammar.[1]

## Introduction

All natural language grammars are ambiguous. Even tightly fitting natural language grammars are ambiguous in some ways. Loosely fitting grammars, which are necessary for handling the variability and complexity of unrestricted text and speech, are worse. The standard technique for dealing with this ambiguity, pruning grammars by hand, is painful, time-consuming, and usually arbitrary. The solution which many people have proposed is to use stochastic models to train statistical grammars automatically from a large corpus.

Attempts in applying statistical techniques to natural language parsing have exhibited varying degrees of success. These successful and unsuccessful attempts have suggested to us that:

- Stochastic techniques combined with traditional linguistic theories *can* (and indeed must) provide a solution to the natural language understanding problem.

- In order for stochastic techniques to be effective, they must be applied with restraint (poor estimates of context are worse than none[8]).

- Interactive, interleaved architectures are preferable to pipeline architectures in NLU systems, because they use more of the available information in the decision-making process.

We have constructed a stochastic parser, $\mathcal{P}$earl, which is based on these ideas.

The development of the $\mathcal{P}$earl parser is an effort to combine the statistical models developed recently into a single tool which incorporates all of these models into the decision-making component of a parser. While we have only attempted to incorporate a few simple statistical models into this parser, $\mathcal{P}$earl is structured in a way which allows any number of syntactic, semantic, and other knowledge sources to contribute to parsing decisions. The current implementation of $\mathcal{P}$earl uses Church's part-of-speech assignment trigram model, a simple probabilistic unknown word model, and a conditional probability model for grammar rules based on part-of-speech trigrams and parent rules.

By combining multiple knowledge sources and using a chart-parsing framework, $\mathcal{P}$earl attempts to handle a number of difficult problems. $\mathcal{P}$earl has the capability to parse word lattices, an ability which is useful in recognizing idioms in text processing, as well as in speech processing. The parser uses probabilistic training from a corpus to disambiguate between grammatically acceptable structures, such as determining prepo-sitional phrase attachment and conjunction scope. Finally, $\mathcal{P}$earl maintains a well-formed substring table within its chart to allow for partial parse retrieval. Partial parses are useful both for error-message generation and for processing ungrammatical or incomplete sentences.

[*]This work was partially supported by DARPA grant No. N0014-85-K0018, ONR contract No. N00014-89-C-0171 by DARPA and AFOSR jointly under grant No. AFOSR-90-0066, and by ARO grant No. DAAL 03-89-C0031 PRI. Special thanks to Carl Weir and Lynette Hirschman at Unisys for their valued input, guidance and support.

[1]The grammar used for our experiments is the string grammar used in Unisys' PUNDIT natural language understanding system.

In preliminary tests, $\mathcal{P}$earl has shown promising results in handling part-of-speech assignment, prepositional phrase attachment, and unknown word categorization. Trained on a corpus of 1100 sentences from the Voyager direction-finding system[2] and using the string grammar from the PUNDIT Language Understanding System, $\mathcal{P}$earl correctly parsed 35 out of 40 or 88% of sentences selected from Voyager sentences not used in the training data. We will describe the details of this experiment later.

In this paper, we will first explain our contribution to the stochastic models which are used in $\mathcal{P}$earl: a context-free grammar with context-sensitive conditional probabilities. Then, we will describe the parser's architecture and the parsing algorithm. Finally, we will give the results of some experiments we performed using $\mathcal{P}$earl which explore its capabilities.

## Using Statistics to Parse

Recent work involving context-free and context-sensitive probabilistic grammars provide little hope for the success of processing unrestricted text using probabilistic techniques. Works by Chitrao and Grishman[3] and by Sharman, Jelinek, and Mercer[14] exhibit accuracy rates lower than 50% using *supervised training*. Supervised training for probabilistic CFGs requires parsed corpora, which is very costly in time and man-power[2].

In our investigations, we have made two observations which attempt to explain the lackluster performance of statistical parsing techniques:

- Simple probabilistic CFGs provide *general* information about how likely a construct is going to appear anywhere in a sample of a language. This average likelihood is often a poor estimate of probability.

- Parsing algorithms which accumulate probabilities of parse theories by simply multiplying them over-penalize infrequent constructs.

$\mathcal{P}$earl avoids the first pitfall by using a context-sensitive conditional probability CFG, where context of a theory is determined by the theories which predicted it and the part-of-speech sequences in the input sentence. To address the second issue, $\mathcal{P}$earl scores each theory by using the geometric mean of the contextual conditional probabilities of all of the theories which have contributed to that theory. This is equivalent to using the sum of the logs of these probabilities.

## CFG with context-sensitive conditional probabilities

In a very large parsed corpus of English text, one finds that the most frequently occurring noun phrase structure in the text is a noun phrase containing a determiner followed by a noun. Simple probabilistic CFGs dictate that, given this information, "determiner noun" should be the most likely interpretation of a noun phrase.

Now, consider only those noun phrases which occur as subjects of a sentence. In a given corpus, you might find that pronouns occur just as frequently as "determiner noun"s in the subject position. This type of information can easily be captured by conditional probabilities.

Finally, assume that the sentence begins with a pronoun followed by a verb. In this case, it is quite clear that, while you can probably concoct a sentence which fits this description and does not have a pronoun for a subject, the first theory which you should pursue is one which makes this hypothesis.

The context-sensitive conditional probabilities which $\mathcal{P}$earl uses take into account the immediate parent of a theory[3] and the part-of-speech trigram centered at the beginning of the theory.

For example, consider the sentence:

> My first love was named $\mathcal{P}$earl.
> (no subliminal propaganda intended)

A theory which tries to interpret "love" as a verb will be scored based on the part-of-speech trigram "adjective verb verb" and the parent theory, probably "S → NP VP." A theory which interprets "love" as a noun will be scored based on the trigram "adjective noun verb." Although lexical probabilities favor "love" as a verb, the conditional probabilities will heavily favor "love" as a noun in this context.[4]

## Using the Geometric Mean of Theory Scores

According to probability theory, the likelihood of two *independent* events occurring at the same time is the product of their individual probabilities. Previous statistical parsing techniques apply this definition to the cooccurrence of two theories in a parse, and claim that the likelihood of the two theories being correct is the product of the probabilities of the two theories.

---

[2]Special thanks to Victor Zue at MIT for the use of the speech data from MIT's Voyager system.

[3]The parent of a theory is defined as a theory with a CF rule which contains the left-hand side of the theory. For instance, if "S → NP VP" and "NP → det n" are two grammar rules, the first rule can be a parent of the second, since the left-hand side of the second "NP" occurs in the right-hand side of the first rule.

[4]In fact, the part-of-speech tagging model which is also used in $\mathcal{P}$earl will heavily favor "love" as a noun. We ignore this behavior to demonstrate the benefits of the trigram conditioning.

This application of probability theory ignores two vital observations about the domain of statistical parsing:

- Two constructs occurring in the same sentence are not necessarily independent (and frequently are not). If the independence assumption is violated, then the product of individual probabilities has no meaning with respect to the joint probability of two events.

- Since statistical parsing suffers from sparse data, probability estimates of low frequency events will usually be inaccurate estimates. Extreme underestimates of the likelihood of low frequency events will produce misleading joint probability estimates.

From these observations, we have determined that estimating joint probabilities of theories using individual probabilities is too difficult with the available data. We have found that the geometric mean of these probability estimates provides an accurate assessment of a theory's viability.

## The Actual Theory Scoring Function

In a departure from standard practice, and perhaps against better judgment, we will include a precise description of the theory scoring function used by $\mathcal{P}$earl. This scoring function tries to solve some of the problems noted in previous attempts at probabilistic parsing[3][14]:

- Theory scores should not depend on the length of the string which the theory spans.

- Sparse data (zero-frequency events) and even zero-probability events do occur, and should not result in zero scoring theories.

- Theory scores should not discriminate against unlikely constructs when the context predicts them.

The raw score of a theory, $\theta$ is calculated by taking the product of the conditional probability of that theory's CFG rule given the context (where context is a part-of-speech trigram and a parent theory's rule) and the score of the trigram:

$$SC_{\text{raw}}(\theta) = \mathcal{P}(rule_\theta|(p_0p_1p_2), rule_{\text{parent}})sc(p_0p_1p_2)$$

Here, the score of a trigram is the product of the mutual information of the part-of-speech trigram,[5] $p_0p_1p_2$, and the lexical probability of the word at the location of $p_1$ being assigned that part-of-speech $p_1$.[6] In the case of ambiguity (part-of-speech ambiguity or multiple parent theories), the maximum value of this product is used. The score of a partial theory or a complete theory is the geometric mean of the raw scores of all of the theories which are contained in that theory.

---

[5]The mutual information of a part-of-speech trigram, $p_0p_1p_2$, is defined to be $\frac{\mathcal{P}(p_0p_1p_2)}{\mathcal{P}(p_0xp_2)\mathcal{P}(p_1)}$, where x is any part-of-speech. See [4] for further explanation.

[6]The trigram scoring function actually used by the parser is somewhat more complicated than this.

**Theory Length Independence** This scoring function, although heuristic in derivation, provides a method for evaluating the value of a theory, regardless of its length. When a rule is first predicted (Earley-style), its score is just its raw score, which represents how much the context predicts it. However, when the parse process hypothesizes interpretations of the sentence which reinforce this theory, the geometric mean of all of the raw scores of the rule's subtree is used, representing the overall likelihood of the theory given the context of the sentence.

**Low-frequency Events** Although some statistical natural language applications employ backing-off estimation techniques[12][5] to handle low-frequency events, $\mathcal{P}$earl uses a very simple estimation technique, reluctantly attributed to Church[8]. This technique estimates the probability of an event by adding 0.5 to every frequency count.[7] Low-scoring theories *will* be predicted by the Earley-style parser. And, if no other hypothesis is suggested, these theories will be pursued. If a high scoring theory advances a theory with a very low raw score, the resulting theory's score will be the geometric mean of all of the raw scores of theories contained in that theory, and thus will be much higher than the low-scoring theory's score.

**Example of Scoring Function** As an example of how the conditional-probability-based scoring function handles ambiguity, consider the sentence

Fruit flies like a banana.

in the domain of insect studies. Lexical probabilities should indicate that the word "flies" is more likely to be a plural noun than an active verb. This information is incorporated in the trigram scores. However, when the interpretation

S → . NP VP

is proposed, two possible NPs will be parsed,

NP → noun (fruit)

and

NP → noun noun (fruit flies).

Since this sentence is syntactically ambiguous, if the first hypothesis is tested first, the parser will interpret this sentence incorrectly.

However, this will not happen in this domain. Since "fruit flies" is a common idiom in insect studies, the score of its trigram, noun noun verb, will be much greater than the score of the trigram, noun verb verb. Thus, not only will the lexical probability of the word "flies/verb" be lower than that of "flies/noun," but also the raw score of "NP → noun (fruit)" will be lower than that of "NP → noun noun (fruit flies)," because of the differential between the trigram scores.

---

[7]We are not deliberately avoiding using all probability estimation techniques, only those backing-off techniques which use independence assumptions that frequently provide misleading information when applied to natural language.

So, "NP → noun noun" will be used first to advance the "S → . NP VP" rule. Further, even if the parser advances both NP hypotheses, the "S → NP . VP" rule using "NP → noun noun" will have a higher score than the "S → NP . VP" rule using "NP → noun."

## Interleaved Architecture in $\mathcal{P}$earl

The interleaved architecture implemented in $\mathcal{P}$earl provides many advantages over the traditional pipeline architecture, but it also introduces certain risks. Decisions about word and part-of-speech ambiguity can be delayed until syntactic processing can disambiguate them. And, using the appropriate score combination functions, the scoring of ambiguous choices can direct the parser towards the most likely interpretation efficiently.

However, with these delayed decisions comes a vastly enlarged search space. The effectiveness of the parser depends on a majority of the theories having very low scores based on either unlikely syntactic structures or low scoring input (such as low scores from a speech recognizer or low lexical probability). In experiments we have performed, this has been the case.

## The Parsing Algorithm

$\mathcal{P}$earl is a time-asynchronous bottom-up chart parser with Earley-type top-down prediction. The significant difference between $\mathcal{P}$earl and non-probabilistic bottom-up parsers is that instead of completely generating all grammatical interpretations of a word string, $\mathcal{P}$earl pursues the N highest-scoring incomplete theories in the chart at each pass. However, $\mathcal{P}$earl parses *without pruning*. Although it is only advancing the N highest-scoring incomplete theories, it retains the lower scoring theories in its agenda. If the higher scoring theories do not generate viable alternatives, the lower scoring theories may be used on subsequent passes.

The parsing algorithm begins with the input word lattice. An $n \times n$ chart is allocated, where $n$ is the length of the longest word string in the lattice. Lexical rules for the input word lattice are inserted into the chart. Using Earley-type prediction, a sentence is predicted at the beginning of the sentence, and all of the theories which are predicted by that initial sentence are inserted into the chart. These incomplete theories are scored according to the context-sensitive conditional probabilities and the trigram part-of-speech model. The incomplete theories are tested in order by score, until N theories are advanced.[8] The resulting advanced theories are scored and predicted for, and the new incomplete predicted theories are scored and added to the chart. This process continues until an complete parse tree is determined, or until the parser decides, heuristically, that it should not continue. The heuristics we used for determining that no parse can be found for an input are based on the highest scoring incomplete theory in the chart, the number of passes the parser has made, and the size of the chart.

## $\mathcal{P}$earl's Capabilities

Besides using statistical methods to guide the parser through the parsing search space, $\mathcal{P}$earl also performs other functions which are crucial to robustly processing unrestricted natural language text and speech.

**Handling Unknown Words** $\mathcal{P}$earl uses a very simple probabilistic unknown word model to hypothesize categories for unknown words. When word which is unknown to the system's lexicon, the word is assumed to be any one of the open class categories. The lexical probability given a category is the probability of that category occurring in the training corpus.

**Idiom Processing and Lattice Parsing** Since the parsing search space can be simplified by recognizing idioms, $\mathcal{P}$earl allows the input string to include idioms that span more than one word in the sentence. This is accomplished by viewing the input sentence as a word lattice instead of a word string. Since idioms tend to be unambiguous with respect to part-of-speech, they are generally favored over processing the individual words that make up the idiom, since the scores of rules containing the words will tend to be less than 1, while a syntactically appropriate, unambiguous idiom will have a score of close to 1.

The ability to parse a sentence with multiple word hypotheses and word boundary hypotheses makes $\mathcal{P}$earl very useful in the domain of spoken language processing. By delaying decisions about word selection but maintaining scoring information from a speech recognizer, the parser can use grammatical information in word selection without slowing the speech recognition process. Because of $\mathcal{P}$earl's interleaved architecture, one could easily incorporate scoring information from a speech recognizer into the set of scoring functions used in the parser. $\mathcal{P}$earl could also provide feedback to the speech recognizer about the grammaticallity of fragment hypotheses to guide the recognizer's search.

**Partial Parses** The main advantage of chart-based parsing over other parsing algorithms is that the parser can also recognize well-formed substrings within the sentence in the course of pursuing a complete parse. $\mathcal{P}$earl takes full advantage of this characteristic. Once $\mathcal{P}$earl is given the input sentence, it awaits instructions as to what type of parse should be attempted for this input. A standard parser automatically attempts to produce a sentence (S) spanning the entire input string. However, if this fails, the semantic interpreter might be able to derive some meaning from the sentence if given non-overlapping noun, verb, and prepositional phrases. If a sentence fails to parse, requests for partial parses of the input string can be made by specifying a range which the parse tree should cover and the category (NP, VP, etc.).

---

[8] We believe that N depends on the perplexity of the grammar used, but for the string grammar used for our experiments we used N=3. For the purposes of training, a higher N should be used in order to generate more parses.

The ability to produce partial parses allows the system to handle multiple sentence inputs. In both speech and text processing, it is difficult to know where the end of a sentence is. For instance, one cannot reliably determine when a speaker terminates a sentence in free speech. And in text processing, abbreviations and quoted expressions produce ambiguity about sentence termination. When this ambiguity exists, $\mathcal{P}$earl can be queried for partial parse trees for the given input, where the goal category is a sentence. Thus, if the word string is actually two complete sentences, the parser can return this information. However, if the word string is only one sentence, then a complete parse tree is returned at little extra cost.

**Trainability** One of the major advantages of the probabilistic parsers is trainability. The conditional probabilities used by $\mathcal{P}$earl are estimated by using frequencies from a large corpus of parsed sentences. The parsed sentences must be parsed using the grammar formalism which the $\mathcal{P}$earl will use.

Assuming the grammar is not recursive in an unconstrained way, the parser can be trained in an unsupervised mode. This is accomplished by running the parser without the scoring functions, and generating many parse trees for each sentence. Previous work[9] has demonstrated that the correct information from these parse trees will be reinforced, while the incorrect substructure will not. Multiple passes of re-training using frequency data from the previous pass should cause the frequency tables to converge to a stable state. This hypothesis has not yet been tested.[10]

An alternative to completely unsupervised training is to take a parsed corpus for any domain of the same language using the same grammar, and use the frequency data from that corpus as the initial training material for the new corpus. This approach should serve only to minimize the number of unsupervised passes required for the frequency data to converge.

## Preliminary Evaluation

While we have not yet done extensive testing of all of the capabilities of $\mathcal{P}$earl, we performed some simple tests to determine if its performance is at least consistent with the premises upon which it is based. The test sentences used for this evaluation are *not* from the training data on which the parser was trained. Using $\mathcal{P}$earl's context-free grammar, these test sentences produced an average of 64 parses per sentence, with some sentences producing over 100 parses.

### Unknown Word Part-of-speech Assignment

To determine how $\mathcal{P}$earl handles unknown words, we removed five words from the lexicon, *i*, *know*, *tee*, *describe*, and *station*, and tried to parse the 40 sample sentences using the simple unknown word model previously described.

In this test, the pronoun, *i*, was assigned the correct part-of-speech 9 of 10 times it occurred in the test sentences. The nouns, *tee* and *station*, were correctly tagged 4 of 5 times. And the verbs, *know* and *describe*, were correctly tagged 3 of 3 times.

| pronoun | 90% |
| --- | --- |
| noun | 80% |
| verb | 100% |
| overall | 89% |

Figure 1: Performance on Unknown Words in Test Sentences

While this accuracy is expected for unknown words in isolation, based on the accuracy of the part-of-speech tagging model, the performance is expected to degrade for sequences of unknown words.

### Prepositional Phrase Attachment

Accurately determining prepositional phrase attachment in general is a difficult and well-documented problem. However, based on experience with several different domains, we have found prepositional phrase attachment to be a domain-specific phenomenon for which training can be very helpful. For instance, in the direction-finding domain, *from* and *to* prepositional phrases generally attach to the preceding verb and not to any noun phrase. This tendency is captured in the training process for $\mathcal{P}$earl and is used to guide the parser to the more likely attachment with respect to the domain. This does not mean that $\mathcal{P}$earl will get the correct parse when the less likely attachment is correct; in fact, $\mathcal{P}$earl will invariably get this case wrong. However, based on the premise that this is the less likely attachment, this will produce more correct analyses than incorrect. And, using a more sophisticated statistical model, this performance can easily be improved.

$\mathcal{P}$earl's performance on prepositional phrase attachment was very high (54/55 or 98.2% correct). The reason the accuracy rate was so high is that the direction-finding domain is very consistent in it's use of individual prepositions. The accuracy rate is not expected to be as high in other domains, although it certainly should be higher than 50% and we would expect it to be greater than 75 %, although we have not performed any rigorous tests on other domains to verify this.

---

[9] This is an unpublished result, reportedly due to Fujisaki at IBM Japan.

[10] In fact, for certain grammars, the frequency tables may not converge at all, or they may converge to zero, with the grammar generating no parses for the entire corpus. This is a worst-case scenario which we do not anticipate happening.

| Preposition | from | to | on | Overall |
|---|---|---|---|---|
| Accuracy Rate | 92 % | 100 % | 100 % | 98.2 % |

Figure 2: Accuracy Rate for Prepositional Phrase Attachment, by Preposition

## Overall Parsing Accuracy

The 40 test sentences were parsed by $\mathcal{P}$earl and the highest scoring parse for each sentence was compared to the correct parse produced by PUNDIT. Of these 40 sentences, $\mathcal{P}$earl produced parse trees for 38 of them, and 35 of these parse trees were equivalent to the correct parse produced by Pundit, for an overall accuracy rate of 88%.

Many of the test sentences were not difficult to parse for existing parsers, but most had *some* grammatical ambiguity which would produce multiple parses. In fact, on 2 of the 3 sentences which were incorrectly parsed, $\mathcal{P}$earl produced the correct parse as well, but the correct parse did not have the highest score.

Of the two sentences which did not parse, one used passive voice, which only occurred in one sentence in the training corpus. While the other sentence,

```
How can I get from cafe sushi to Cambridge
City Hospital by walking
```

did not produce a parse for the entire word string, it could be processed using $\mathcal{P}$earl's partial parsing capability. By accessing the chart produced by the failed parse attempt, the parser can find a parsed sentence containing the first eleven words, and a prepositional phrase containing the final two words. This information could be used to interpret the sentence properly.

## Future Work

The $\mathcal{P}$earl parser takes advantage of domain-dependent information to select the most appropriate interpretation of an input. However, the statistical measure used to disambiguate these interpretations is sensitive to certain attributes of the grammatical formalism used, as well as to the part-of-speech categories used to label lexical entries. All of the experiments performed on $\mathcal{P}$earl thus far have been using one grammar, one part-of-speech tag set, and one domain (because of availability constraints). Future experiments are planned to evaluate $\mathcal{P}$earl's performance on different domains, as well as on a general corpus of English, and on different grammars, including a grammar derived from a manually parsed corpus.

Future work should also investigate $\mathcal{P}$earl's performance on speech data. By incorporating the speech recognizer's acoustic score into the parser's scoring function, one could investigate the parser's ability to select the appropriate word strings from an N-best list of a speech recognizer's output.

## Conclusion

The probabilistic parser which we have described provides a platform for exploiting the useful information made available by statistical models in a manner which is consistent with existing grammar formalisms and parser designs. $\mathcal{P}$earl can be trained to use any context-free grammar, accompanied by the appropriate training material. And, the parsing algorithm is very similar to a standard bottom-up algorithm, with the exception of using theory scores to order the search.

More thorough testing is necessary to measure $\mathcal{P}$earl's performance in terms of parsing accuracy, part-of-speech assignment, unknown word categorization, idiom processing capabilities, and even word selection in speech processing. With the exception of word selection, preliminary tests show $\mathcal{P}$earl performs these tasks with a high degree of accuracy. But, in the absence of precise performance estimates, we still propose that the architecture of this parser is preferable to traditional pipeline architectures. Only by using an interleaved architecture can a speech recognizer efficiently make use of complex grammatical information to select from among hypothesized words.

## References

[1] Ayuso, D., Bobrow, R., et. al. 1990. Towards Understanding Text with a Very Large Vocabulary. In Proceedings of the June 1990 DARPA Speech and Natural Language Workshop. Hidden Valley, Pennsylvania.

[2] Brill, E., Magerman, D., Marcus, M., and Santorini, B. 1990. Deducing Linguistic Structure from the Statistics of Large Corpora. In Proceedings of the June 1990 DARPA Speech and Natural Language Workshop. Hidden Valley, Pennsylvania.

[3] Chitrao, M. and Grishman, R. 1990. Statistical Parsing of Messages. In Proceedings of the June 1990 DARPA Speech and Natural Language Workshop. Hidden Valley, Pennsylvania.

[4] Church, K. 1988. A Stochastic Parts Program and Noun Phrase Parser for Unrestricted Text. In Proceedings of the Second Conference on Applied Natural Language Processing. Austin, Texas.

[5] Church, K. and Gale, W. 1990. Enhanced Good-Turing and Cat-Cal: Two New Methods for Estimating Probabilities of English Bigrams. *Computers, Speech and Language.*

[6] Church, K. and Hanks, P. 1989. Word Association Norms, Mutual Information, and Lexicography. In Proceedings of the 27th Annual Conference of the Association of Computational Linguistics.

[7] Fano, R. 1961. *Transmission of Information.* New York, New York: MIT Press.

[8] Gale, W. A. and Church, K. 1990. Poor Estimates of Context are Worse than None. In Proceedings of the June 1990 DARPA Speech and Natural Language Workshop. Hidden Valley, Pennsylvania.

[9] Hindle, D. 1988. Acquiring a Noun Classification from Predicate-Argument Structures. Bell Laboratories.

[10] Hindle, D. and Rooth, M. 1990. Structural Ambiguity and Lexical Relations. In Proceedings of the June 1990 DARPA Speech and Natural Language Workshop. Hidden Valley, Pennsylvania.

[11] Jelinek, F. 1985. Self-organizing Language Modeling for Speech Recognition. IBM Report.

[12] Katz, S. M. 1987. Estimation of Probabilities from Sparse Data for the Language Model Component of a Speech Recognizer. *IEEE Transactions on Acoustics, Speech, and Signal Processing, Vol. ASSP-35, No. 3.*

[13] Seneff, Stephanie 1989. TINA. In Proceedings of the August 1989 International Workshop in Parsing Technologies. Pittsburgh, Pennsylvania.

[14] Sharman, R. A., Jelinek, F., and Mercer, R. 1990. In Proceedings of the June 1990 DARPA Speech and Natural Language Workshop. Hidden Valley, Pennsylvania.